# PATH PLANNER FOR OBJECTS, ROBOTS AND MANNEQUINS BY MULTI-AGENTS SYSTEMS OR MOTION CAPTURES

**D. Chablat**

Institut de Recherche en Communications et Cybernétique de Nantes (UMR CNRS 6597)

École Centrale de Nantes, 1, rue de la Noë, 44321 Nantes

Damien.Chablat@irccyn.ec-nantes.fr

**ABSTRACT**

In order to optimise the costs and time of design of the new products while improving their quality, concurrent engineering is based on the digital model of these products. However, in order to be able to avoid definitively physical model without loss of information, new tools must be available. Especially, a tool making it possible to check simply and quickly the maintainability of complex mechanical sets using the numerical model is necessary. Since one decade, the MCM team of IRCCyN works on the creation of tools for the generation and the analysis of trajectories of virtual mannequins. The simulation of human tasks can be carried out either by robot-like simulation or by simulation by motion capture. This paper presents some results on the both two methods. The first method is based on a multi-agent system and on a digital mock-up technology, to assess an efficient path planner for a manikin or a robot for access and visibility task taking into account ergonomic constraints or joint limits. The human operator is integrated in the process optimisation to contribute to a global perception of the environment. This operator cooperates, in real-time, with several automatic local elementary agents. In the second method, we worked with the CEA and EADS/CCR to solve the constraints related to the evolution of human virtual in its environment on the basis of data resulting from motion capture system. An approach using of the virtual guides was developed to allow to the user the realization of precise trajectory in absence of force feedback.

**KEYWORDS**

Multi-agent system, Ergonomics, Motion planner, Motion capture, Virtual guide.

## 1. INTRODUCTION

In an industrial environment, the access to a sharable and global view of the enterprise project, product, and/or service appears to be a key factor of success. It improves the triptych delay-quality-cost but also the communication between the different partners and their implication in the project. For these reasons, the digital mock-up (DMU) and its functions are deeply investigated by industrials. Based on computer technology and virtual reality, the DMU consists in a platform of visualization and simulation that can cover different processes and areas during the product lifecycle such as product design, industrialization, production, maintenance, recycling and/or customer support. The digital model enables the earlier identification of possible issues and a better understanding of the processes even, and maybe above all, for actors who are not specialists. Thus, a digital model allows deciding before expensive physical prototypes have been built. Even if evident progresses were noticed and applied in the domain of DMUs, significant progresses are still awaited for a placement in an industrial context. As a matter of fact, the digital model offers a way to explore areas such as maintenance or ergonomics of the product that were traditionally ignored at the beginning phases of a project; new processes must consequently be developed.

Through the integration of a manikin or a robot in a virtual environment, the suitability of a product, its shape and functions can be assessed. In the same time, it becomes possible to settle the process for assembling with a robot the different components of the product. Moreover, when simulating a task that should be performed by an operator with a virtual manikin model, feasibility, access and visibility can be checked. The conditions of the performances in terms of efforts, constraints and comfort can also be analysed. Modifications on the process, on the product or on the task itself may follow but also a better and earlier training of the operators to enhance their performances in the real environment. Moreover, such a use of the DMU leads to a better conformance to health and safety standards, to a maximization of human comfort and safety and an optimisation of the robot abilities.

With virtual reality tools such as 3D manipulators, it is possible to manipulate the object as easily as in a real to manipulate the object as easily as in a real environment. Some drawbacks are the difficulty to manipulate the object with as ease as in a real environment, due to the lack of kinematics constraints and the automatic collision avoidance. As a matter of fact, interference detection between parts is often displayed through color changes of parts in collision but collision is not avoided.

Another approach consists in integrating automatic functionality into the virtual environment in order to ease the user's task. Many research topics in the framework of robotics dealing with the definition of collision-free trajectories for solid objects are also valid in the DMU. Some methodologies need a global perception of the environment, like (i) visibility graphs proposed by Lozano-Pérez and Wesley [1], (ii) geodesic graphs proposed by Tournassoud [2], or (iii) Voronoï's diagrams [3]. However, these techniques are very CPU consuming but lead to a solution if it exists. Some other methodologies consider the moves of the object only in its close or local environment. The success of these methods is not guaranteed due to the existence of local minima. A specific method was proposed by Khatib [4] and enhanced by Barraquand and Latombe [5]. In this method, Khatib's potentials method is coupled with an optimization method that minimizes the distance to the target and avoids collisions. All these techniques are limited, either by the computation cost, or the existence of local minima as explained by Namgung [6]. For these reasons a designer, is required in order to validate one of the different paths found or to avoid local minima.

The accessibility and the optimum placement of an operator to perform a task is also a matter of path planning that we propose to solve with DMU. In order to shorten time for a trajectory search, to avoid local minima and to suppress tiresome on-line manipulation, we intend to settle for a mixed approach of the above presented methodologies. Thus, we use local algorithm abilities and global view ability of a human operator, with the same approach as [7]. Among the local algorithms, we present these ones contributing to a better visibility of the task, in term of access but also in term of comfort.

## 2. PATH PLANNING AND MULTI-AGENT ARCHITECTURE

The above chapter points out the local abilities of several path planners. Furthermore, human global vision can lead to a coherent partition of the path planning issue. We intend to manage simultaneously these local and global abilities by building an interaction between human and algorithms in order to have an efficient path planner [8] for a manikin or a robot with respect of ergonomic constraints or joints and mechanical limits of the robot.

### 2.1 HISTORY

Several studies about co-operation between algorithm processes and human operators have shown the great potential of co-operation between agents. First concepts were proposed by Ferber [9]. These studies led to the creation of a "Concurrent Engineering" methodology based on network principles, interacting with cells or modules that represent skills, rules or workgroups. Such studies can be linked to work done by Arcand and Pelletier [10] for the design of a cognition based multi-agent architecture. This work presents a multi-agent architecture with human and society behaviour. It uses cognitive psychology results within a co-operative human and computer system. All these studies show the important potential of multi-agent systems (MAS). Consequently, we built a "positioning" manikin, based on MAS, that combines human interactive integration and algorithms.

### 2.2 CHOICE OF THE MULTI-AGENT ARCHITECTURE

Several workgroups have established rules for the definition of the agents and their interactions, even for dynamic architectures according to the environment evolution [9, 11]. From these analyses, we keep the following points for an elementary agent definition. An elementary agent: (i) is able to act in a common environment, (ii) is driven by a set of tendencies (goal, satisfaction function, etc.), (iii) has its own resources, (iv) can see locally its environment, (v) has a partial representation of the environment, (vi) has some skills and offers some services, and (vii) has behavior in order to satisfy its goal, taking into account its resources and abilities, according to its environment analysis and to the information it receives.

The points above show that direct communications between agents are not considered. In fact, our architecture implies that each agent acts on its set of variables from the environment according to its goal. Our Multi Agent System (MAS) will be a blackboard-based architecture.

## 2.3 PATH PLANNING AND MAS

The method used in automatic path planners is schematised Figure 1a. A human global vision can lead to a coherent partition of the main trajectory as suggested in [12]. Consequently, another method is the integration of an operator to manage the evolution of the variables, taking into account his or her global perception of the environment (Figure 1b). To enhance path planning, a coupled approach using multi-agent and distributed principles as it is defined in [8] can be build; this approach manages simultaneously the two, local and global, abilities as suggested Figure 1c. The virtual site enables graphic visualization of the database for the human operator, and communicates positions of the virtual objects to external processes.

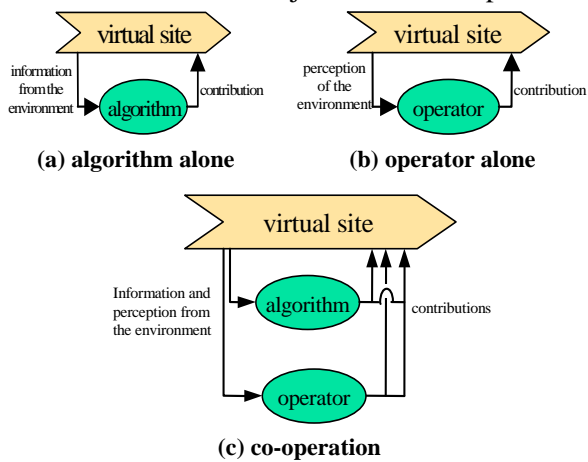

**Figure 1. Co-operation principles.**

As a matter of fact, this last scheme is clearly correlated with the blackboard based MAS architecture. This principle is described in [9, 13, 11]. A schematic presentation is presented on Figure 2. The only medium between agents is the common database of the virtual reality environment. The human operator can be considered as an elementary agent for the system, co-operating with some other elementary agents that are simple algorithms.

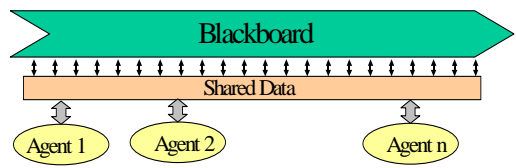

**Figure 2. Blackboard principle with co-operating agents.**

## 2.4 CONSIDERED APPROACH

The approach we retained is the one proposed in [7] whose purpose was to validate new CAD/CAM solutions based on a distributed approach using a virtual reality environment. This method has successfully shown its advantage by demonstrating in a realistic time the assembly task of several components with a manikin. Such problem was previously solved by using real and physical mock-ups. We kept the same architecture and developed some elementary agents for the manikin (Figure 3). In fact, each agent can be recursively divided in elementary agents.

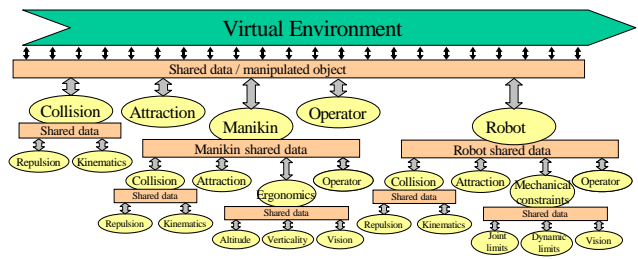

**Figure 3. Co-operating agents and path planning activity [7].**

Each agent $i$ acts with a specific time sampling which is pre-defined by a specific rate of activity $\lambda_i$. When acting, the agent sends a contribution, normalized by a value $\Delta_i$, to the environment and/or the manipulated object (the manikin in our study). In Figure **4**, we represent the Collision agent with a rate of activity equal to 1, the Attraction agent has a rate of 3 and Operator and Manikin agents a rate of 9. This periodicity of the agent actions is a characteristic of the architecture: it expresses a priority between each of the goals of the agents. To supervise each agent activity, we use an event programming method where the main process collects agent contributions and updates the database [7]. The normalization of the actions of the agents (the values $\Delta_i$) induces that the actions are relative and not absolute.

## 2.5 EXAMPLES

The former method is illustrated with two different examples. The first one (Figure 4) uses the MAS for testing the ability of a manikin for mounting an oxygen bottle inside an airplane cockpit through a trap. During the path planning process, the operator has acted in order to drive the oxygen bottle toward the middle of the trap. The other agents have acted in order to avoid collisions and to attract the oxygen bottle toward the final location. The real time duration is approximately 30s. The number of degrees of freedom is equal to 23. The second example (Figure 5) is related to the automatic manipulation of a robot which base is attracted toward a wall. The joints are managed by the agents in order to avoid a collision and to solve the associated inverse kinematic model.

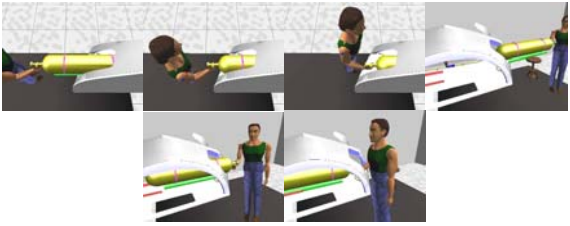

**Figure 4 Trajectory path planning of a manikin using the MAS**

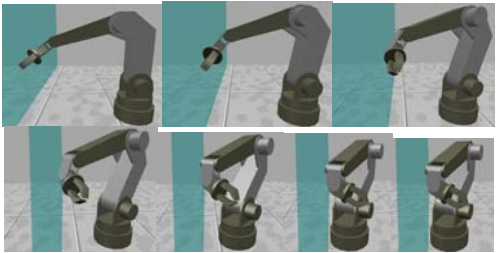

**Figure 5 Trajectory path planning of a robot using the MAS**

## 3. VISIBILITY AND MAINTAINABILITY CHECK WITH MULTI-AGENT SYSTEM IN VIRTUAL REALITY

### 3.1 INTRODUCTION

For the visibility check, we have focused our attention on the trunk and the head configurations of a manikin (resp. the end-effector and the film camera orientation of a robot). The joint between the head and the trunk is characterized by three rotations $\alpha_b$, $\beta_b$ and $\theta_b$ whose range limits are defined by ergonomic constraints (Figure 6) (resp. the joint limits of the robot).

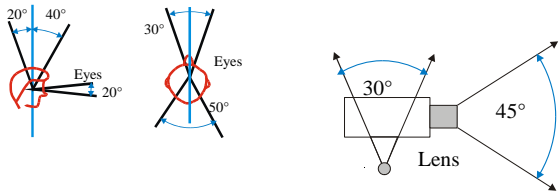

**Figure 6. Example of joint limits and visibility capacity of a manikin and of a film camera.**

These data can be found using the results of ergonomic research [14]. To solve the problem of visibility, we define a cone C whose vertex is centered between the two eyes (resp. the center of the film camera) and whose base is located in the plane orthogonal to **u**, centered on the target (Figure 7). The cone width $\varepsilon_c$ is variable.

Thus, additionally to the position and orientation variables of all parts in the cluttered environment (including the manikin itself), we consider in particular:

• Three degrees of freedom for the manikin (resp. the robot) to move it in the x-y plane: $\mathbf{x}_m = (x_m, y_m, \theta_m,)^T$. It is also possible to take into account a degree of freedom $z_m$ if we want to give to the manikin (resp. the robot) the capacity to clear an obstacle.

• Three degrees of freedom for the head joint (resp. wrist joints) to manage the manikin (resp. robot) vision: $\mathbf{q}_b = (\alpha_b, \beta_b, \theta_b)^t$ with their corresponding joint constraints.

The normalized contributions from the agents are defined with two fixed parameters: $\Delta_{pos}$ for translating moves and $\Delta_{or}$ for rotating moves.

### 3.2 AGENTS ENSURING VISIBILITY ACCESS AND COMFORT FOR MANIKIN, AND VISIBILITY ACCESS FOR ROBOT

We present below all the elementary agents used in our system to solve the access and visibility task.

• *Attraction* agent for the manikin (resp. the robot)
The goal of the *attraction* agent is to enable the manikin (resp. the robot) to reach the target with the best trunk posture (resp. the best base placement of the robot), that is:

• To orient the projection of $\mathbf{y}_m$ on the floor plane collinear to the projection of **u** on the same plane by rotation of $\theta_m$ (Figure **7**),

• To position $x_m$ and $y_m$, coordinates of the manikin (resp. the robot) in the environment floor, as close as possible to the target position (Figure **7**), and for the robot.

• $q_1$ up to $q_n$ using the inverse kinematic model. This last agent acts in order to keep the robot posture, as much as possible, in the same aspect, or posture, or configuration as defined in [15].)

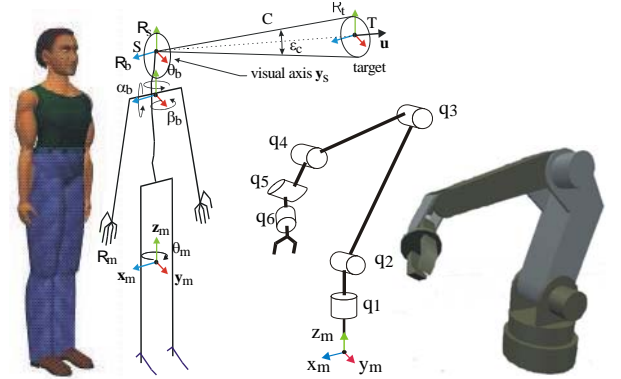

**Figure 7. Manikin skeleton and robot kinematics; visibility cone and target definition.**

This *attraction* agent only considers the target and does not take care of the environment. This agent is similar to the attraction force introduced by Khatib

[4], and gives the required contributions $x_{att}$, $y_{att}$, and $\theta_{att}$ according to the attraction toward a target referenced as above. These contributions, which act on the manikin (resp. the robot) leading member position and orientation (in our case the trunk (resp. the base of the robot and its kinematics)), are normalized according to $\Delta_{pos}$ and $\Delta_{or}$.

➢ *Repulsion agent between manikin (resp. robot) and the cluttered environment.*

This *repulsion* agent acts in order to avoid the collisions between the manikin (resp. the robot) and the cluttered environment, which may be static or mobile.

Several possibilities can be used in order to build a collision criterion. The intersection between two parts A and B in collision, as shown by Figure 8a, can be quantified in several ways. We can consider either the volume V of collision, or the surface Σ of collision, or the depth $D_{max}$ of collision (Figure 8b). The main drawback of these approaches comes out from the difficulty to determine these values. Moreover, 3D topological operations are not easy because many of the virtual reality softwares use polyhedral surfaces to define 3D objects. To determine $D_{move}$, the distance to avoid the collision (Figure 8b), we have to store former positions of the mobile (manikin or robot), so this quantification does not use only the database at a given instant but uses former information. This solution cannot be kept with our blackboard architecture that only provides global environment status at an instant.

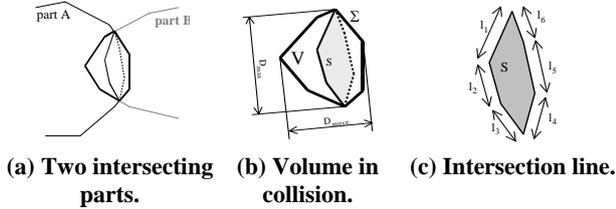

**(a) Two intersecting parts.** **(b) Volume in collision.** **(c) Intersection line.**

**Figure 8 Collision criteria.**

Another quantification of the collision is possible with the use of the collision line between the two parts. With this collision line, we can determine the maximum surface S or the maximum length of the collision line $l = \Sigma\ l_i$ (Figure 8c). By the end, we compute the gradient of the collision criterion according to the Cartesian environment frame using a finite difference approximation.

From the gradient vector of the collision length $\mathbf{grad}_{(x,y,\theta)}(l)$, contributions $x_{rep}$, $y_{rep}$, and $\theta_{rep}$ are computed by the *repulsion* agent. These contributions, acting on the manikin trunk position and orientation, are normalized according to $\Delta_{pos}$ and $\Delta_{or}$.

➢ *Head orientation agent*

The goal of the *head orientation* agent is to rotate the head of the manikin (resp. of the film camera) in order to observe the target. It ensures the optimum configuration that maximizes visual comfort (resp. the visibility of the target). Finding the optimum configuration consists in minimizing efforts on the joint coupling the head with the trunk and minimizing ocular efforts (resp. mechanical efforts or isotropy of the configuration). We simplify the problem by considering that the manikin has a monocular vision, defined by a cone whose principal axis, called vision axis, is along $\mathbf{y}_s$ and whose vertex is the center of manikin eyes (in that case, the manikin vision is similar to that of a film camera on a robot). If the target belongs to the vision axis, ocular efforts are considered null. Our purpose consists in orienting $\mathbf{y}_s$ such as it becomes collinear to $\mathbf{u}$ by rotation of $\alpha_b$ and $\theta_b$ (Figure **7**), subject to joint limits. A joint limit average for an adult is given in Figure 6. In the case of a film camera, the corresponding values will be the optical characteristics of the film camera.

The algorithm of this agent is similar to the *attraction agent* algorithm presented there above; contributions $\alpha_{head}$ and $\theta_{head}$, after normalization, are applied to the joint coupling the head to the manikin trunk (resp. the wrist joints of the robot).

➢ Visibility *agent*

The *visibility* agent ensures that the target is visible, that is, no interference occurs between the segment ST linking the center of manikin eyes (resp. of the film camera) and the target, and the cluttered environment. The repulsion algorithm is exactly the same as the one presented there above:

• we determine the collision line length,

• if non equal to zero, normalized contributions are determined from $x_{vis}$, $y_{vis}$, and $\theta_{vis}$ computed by the *visibility* agent according to the gradient vector of the collision length,

• contributions are applied to the manikin trunk (resp. the base of the robot).

It is to notice that some contributions may also be applied to the head orientation (resp. the film camera orientation). This is due to the fact that by turning the head, collisions between the simplified cone with the environment may also occur.

The use of a simplified cone offers the advantage of combining an ergonomic criterion with the repulsion effect. As a matter of fact, when the vision axis $\mathbf{y}_s$ is inside the cone C (Figure **7**), we widen the cone, respecting a maximum limit. If not, we decrease its vertex angle, also with respect of a minimum limit that corresponds to the initial condition when starting this *visibility* agent. The maximum limit may be expressed according to the target size or/and to the type of task to perform: proximal or distant visual checking, global or specific area to control.

➢ Operator *agent on the manikin (resp. on the robot)*

One of the aims of the study is to integrate a human operator within the MAS in order to operate in real-time. The *operator* has a global view of the cluttered environment displayed by means of the virtual reality software. Her or his action must be simple and efficient. For that purpose, we use a Logitech SpaceMouse device that allows us to manipulate a body with six degrees of freedom.

The action of the *operator* agent only considers the move of the leading object, which is in our case the manikin trunk (resp. the base of the robot). Parameters come out from position $x_{op}$ and $y_{op}$ and orientation $\theta_{op}$ returned from the SpaceMouse. These contributions are normalized, in the same way as with the *attraction* or *repulsion* agents.

## 3.2 RESULTS AND CONCLUSIONS

This method has been tested to check the visual accessibility of specific elements under a trap of an aircraft. The digital model is presented in Figure 9 and the list of elementary agents is depicted in the master agent window in Figure 10. In this example, the *repulsion* agent for the manikin (**Repulsion**), the *visibility* agent (**Visual**) and the *head orientation* agent (**Cone**) have a specific rate of activity equal to 1, meaning that their actions have priority but it is possible for the operator to change in real-time this activity rate. Since the action of each agent is independent from the other elementary agents, it is possible to inactivate some of them (**Pause/Work** buttons). The values of $\Delta_{pos}$ and $\Delta_{or}$, which are used to normalize the agent contributions, can also be modified in real-time (**Position** and **Orientation** buttons) in order to adapt the contribution to the scale of the environment or to the task to perform. Our experience shows that the contribution of the human operator is important in the optimization process. Indeed, if the automatic agent process fails (which can be the case when the cone used in the *visibility* agent is in collision with the environment), the human operator can:

• give to the MAS intermediate targets that will lead to a valid solution;

• move the manikin to a place where the MAS process could find a solution.

On the other hand, the MAS allow the human operator to act more quickly and more easily with the DMU. The elementary agents guarantee a good physical and visual comfort and enable to quantify and qualify it, which would be a hard task for the human operator, even with sophisticated virtual reality devices. For instance, we can evaluate the rotations of the head and see how they are dispersed from a neutral configuration, inducing little effort. Moreover, the MAS permits us a very good local collision detection and avoidance without any effort of the human operator.

The advantage of the MAS is to enable the combination of independent elementary agents to solve complex tasks. Thus, the agents participating in the visibility task can be coupled with agents enabling accessibility and maintainability as proposed by Chedmail [7]. The purpose of further work consists in a global coupling of manikin manipulations taking into account visual and ergonomic constraints and the manipulation of moving objects as robots. The result of this work was implemented in an industrial context with Snecma Motors [16].

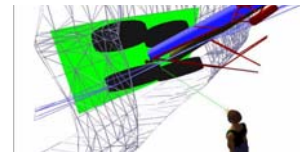
**Figure 9. Digital model of a trap of an aircraft.**

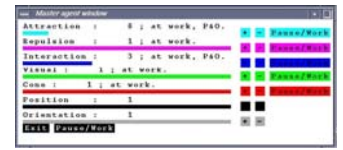
**Figure 10. Master agent window.**

## 4. PATH PLANNING AND MOTION CAPTURE

Now, we focus our work on interactive animation that makes it able to drive an avatar in Real Time (RT) through a motion capture device, as seen on Figure 11 [29]. In our framework, we would like to be able to drive, in a "realistic way", a manikin, doing sharp tasks such as the ones that can be done by a worker, e.g. screwing a screw, sawing, drilling a hole, or nailing down a nail. These tasks all need sharp movements of the worker in the real world, and need the same accuracy in a virtual world. Thus to be able to be "realistic", we will have to emulate the real world's physical laws, and human specificities: interaction with environment, non-penetration of objects, joint limits enforcement, human-like motion of the avatar.

The main point in the features we want to be implemented is *interaction with environment*, because it drives the choice of the model we are to put into action. If we want to be able to interact in a natural way, we have to implement natural behaviours. That is interaction must be done through forces. This means kinematical approaches are not adequate: we must use the equations of dynamics. This perfectly fits the general context of our work: our architecture will be coupled to a portable haptic device being developed in [25]. The system we aim at controlling is complex. An effective approach in such cases is to implement a modular architecture, so as to decouple functionalities of our systems in different modules. This will be made possible thanks to a passive approach. Actually, passivity will allow us to bring the modularity a step further than the simple functional modularity. Indeed it will allow decoupling the analysis of the stability of the whole system, into the analysis of the passivity of each module.

The approaches enabling tasks prioritisation such as [21] seem interesting in case of conflicting tasks. Unfortunately, they cannot be used in the context of passivity, because they use projections. As we will show, in the general case, the use of projections while optimising other potentials is not compatible with passivity: in general projections break passivity. This could seem very disturbing. Nevertheless, this limit only appears in case of impossibility for the avatar to reproduce the movements of the actor. In the context of engineering (at least), one does not really want to control the virtual human in the case of unfeasible movements. Indeed what we really want is to know if the movement is feasible or not, this makes a big difference.

Knowing this, we propose other control modes, which help the manikin achieving its movements, instead of trying to control it performing movements it will never succeed in. These control modes are relevant because of a loss of information when we are immersed in simulations. Haptic devices' approach is interesting, but requires an heavy infrastructure. Thus, if we wanted to be able to perform precision control, with a light infrastructure we would use *virtual guides*, based on projections. Nevertheless, as explained above, we will show that applying a projection "as is" can lead to the loss of the passive nature of our controller, so we propose a way to build *physical projections*, which respect passivity, thanks to mechanical analogies.

We also implement a solution to solve for contacts. Zordan, and Hodgins [23], propose a solution that "hits and reacts", modifying control gains during the simulation. This leads to an approach that is known to be unsafe because of instabilities, and is not real-time. Schmidl, and Lin [24] use a hybrid they call geometry-driven physics that uses IK to solve for the manikin's reaction to contact, and impulse-based physics for the environment. They lose the physical nature of the simulation. The method we use does not make these concessions. The global control scheme we want is depicted in Figure 11 . We know motion capture positions are the inputs of a controller (which is about to be detailed), this controller drives a physical simulation (also to be detailed); which sends the updated world configuration to the output renderer. The manikin we aim at controlling is composed of two layers: a skeleton (which can be viewed as a kinematical chain), and a rigid skin on top of it, which will be useful for collision detection (of course, once motion is calculated, it can be sent to a nice renderer which will tackle with soft deforming skin – which is out of scope). The scheme of Figure 12, shows the whole architecture.

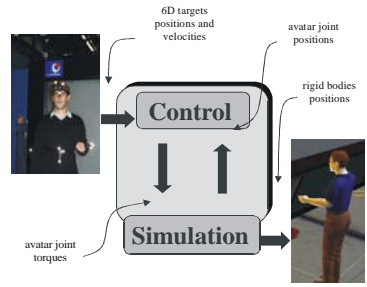 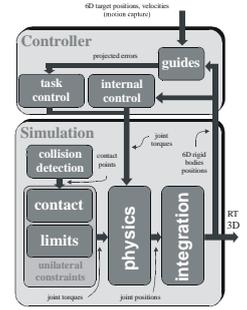

**Figure 11 Global scheme of the system**    **Figure 12 Detailed global control scheme**

## 4.1 SIMULATION

The blocks *physics*, and *integration* of the scheme Figure 12 should be self-motivated, as we want to emulate the physical laws of the real world. Let's just describe some choices we made.

**Physics:** As stated above, the simulation is to support contact, and interaction with the environment, that is forces. Hence dynamics established itself as the best choice for our model. In a first attempt, we have used a first order dynamical model, as stated in [19] (though simpler it highlights the problem to be solved). Integration is done through an additional joint damping term, that is

$$A(q)\ddot{q} + C(q,\dot{q})\dot{q} + G(q) + B_a\dot{q} = \Gamma \text{ or } B_a\dot{q} = \Gamma \quad (1)$$

with $B_a$, the damping matrix chosen to be symmetric positive semi-definite; $q$, and $\Gamma$ are the joint parameters and torques.

Of course, these physical laws are intrinsically passive at port $<\dot{q},\Gamma>$, the proof can be found in [17].

**Integration:** We use a Runge-Kutta-Munthe-Kaas scheme (see [20]). This is a Runge-Kutta method dedicated to integration on Lie Groups, that is known to be efficient. This work is left to Generalized Virtual Mechanisms (GVM) [27], a library being developed in CEA\LIST.

## 4.2 CONTROL

In order to understand the relevance of the other functional blocks of the diagram, we have to wonder *what makes a human move, and have the specific motion he has*? In doing so, we found three sources of movement (or influential factors):

• The first one enables an end-effector to reach a goal in task space. (Ex: I want my left hand to reach a plug on the wall)

• The second one drives configuration, or gaits. (Ex: it makes the difference between the gait of a

fashion model, and the gait of an old cowboy)

- The last one enforces physical, and biomechanical constraints. (Ex: joint limits, non-penetration with environment, but also balance control, which will be taken into account in a forthcoming paper as in [28])

These influential factors can be translated straightforwardly into control idioms, being task space, and null-space control under unilateral constraints. Bilateral and unilateral constraints will not cause any problem. But operational space, and null-space control must be studied carefully if one does not want to break passivity.

1) External task space control

At first glance it could seem interesting to bring prioritisation between external tasks. Nevertheless, we show that such prioritisations can break passivity. This urges us to use other control modes. In [21], Sentis, and Khatib introduce dynamical decoupling of $n$ external tasks. The joint torques $\Gamma$ they apply on their manikin is composed of the influence $\Gamma_i$ of $n$ prioritised external tasks, such that lower priority tasks do not disturb higher ones:

$$\Gamma = \sum_{i=1}^{n} \Gamma_{i|prev(i)}, \text{ with } \Gamma_{i|prev(i)} = \Pi_{prev(i)}^T \Gamma_i, \quad (2)$$

with $\Pi_{prev(i)}^T$ projecting into higher priority tasks' null-space. Using such projections is unsafe; we show they can break passivity. Let's take an example with two external ports $(J_1, W_1, V_1, \Gamma_1)$, and $(J_2, W_2, V_2, \Gamma_2)$, each port being described by its Jacobian, the wrench applied, its velocity, and the torques it generates. We give task 1 the highest priority, and $\Pi_1$ is the projection allowing enforcing priorities. The priority appears if $\Pi_1^T$ projects into $Ker(J_1 B_a^{-1})$. For the passivity to be ensured, we must enforce:

$$\int_0^t (W_1^T V_1 + W_2^T V_2) dt \geq -\beta^2, \text{ with } \beta \text{ real}, \forall W_1, \text{ and } W_2 \quad (3)$$

There always exists $W_1$, $W_2$, and $J_2$, such that:

$$\begin{cases} J_2^T W_2 \neq 0 \\ J_1^T W_1 + \frac{1}{2} J_2^T W_2 = 0, \\ J_2 B_a^{-1} \Pi_1^T J_2^T = 0 \end{cases} \quad (4)$$

Thus in the general case, projecting external interactions can break passivity. As explained earlier, projections are useful in case of conflicting targets, in the case when the manikin cannot achieve what it is asked to do. It means that a real human with the same morphology as its virtual counterpart, could neither achieve the movement. In the case of engineering, we do not look for controlling virtual humans, in cases where they cannot achieve the target motion. We only want to know if the movement is feasible or not. Thus the proposed control is rather simple, but behaves well in case of unfeasible movements, and is able to warn in case of infeasibility. In [26], they use a 6D Proportional Derivative (PD) operational space controller at each point to be controlled on the manikin (Figure 13):

$$f_{ctrl} = K(x_d - x) + B_c(v_d - v). \quad (5)$$

Note that if the control points where linked to their targets position, thanks to a damped spring, the force generated would have the same shape; this makes it able to "draw" the controller, as its mechanical analogy, the damped spring.

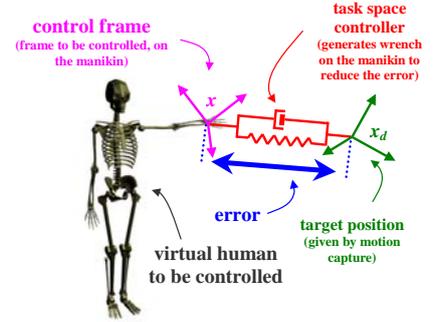

**Figure 13 Task space control**

Concatenating (1), and (5), we obtain

$$B_a \dot{q} = J^T (K(x_d - x) + B_c(v_d - v)), \text{ and}$$

$$(B_a + J^T B_c J)\dot{q} = J^T (K(x_d - x) + B_c v_d).$$

This equation admits a solution if $(B_a + J^T B_c J)$ is not singular, that is $B_a$ must be definite, or $J$ must be full rank, with $B_c$ definite, then:

$$\dot{q} = (B_a + J^T B_c J)^{-1} J^T (K(x_d - x) + B_c v_d). \quad (6)$$

2) Constraints

All the constraints enumerated above (joint limits, contact…) are unilateral. They can all be solved through Linear Complementary Problems (LCP) algorithms. We use GVM's unilateral constraints solver [27, 29]. Using an approach similar to Ruspini, and Khatib [18], we express the contact problem in an LCP form, which is solved for $f$ - the contact wrench. This stage is passive so long as the dynamical equation is passive.

3) Configurations (internal control)

Null space control is usually solved as in [22], optimizing internal potentials (the choice of these potentials is out of scope). If we do not want this optimization to disturb external control, we must work in the null space of our external task, projecting the joint torques induced from the internal potential. Here we show this projection can break passivity. Let's take a skeleton, an internal potential $U(q)$, to be minimized (its associated joint torques are $\Gamma_{int}$), and an external port (defined as in 1) by $(J_1, W_1, V_1, \Gamma_1)$). Projection $\Pi_1$ (to be defined) is to give priority to the external task:

$$\Gamma_{int} = -\alpha \Pi_1^T \frac{\partial U}{\partial q}.$$

Speed at the external port 1 is:

$$V_1 = J_1 \dot{q} = J_1 B^{-1} \Gamma = J_1 B_a^{-1} \left( J_1^T W_1 - \alpha \Pi_1^T \frac{\partial U}{\partial q} \right), \quad (7)$$

If one does not want the internal potential to disturb the external, we must take $\Pi_1^T$ as a projection in $Ker(J_1 B_a^{-1})$. We must ensure this projection keeps passivity at all ports. At the external port, thanks to (7) we can write:

$$W_1^T V_1 = W_1^T J_1 B_a^{-1} J_1^T W_1 \geq 0. \quad (8)$$

So the system composed of the external task coupled to the internal one is passive at the external port, if the external port was passive before coupling. The internal potential's influence does not disappear as in (8). Then, projecting an internal potential can break passivity. Nevertheless, there are solutions to this problem:

- We can reduce the internal task's influence through $\alpha$ such that the system remains passive.

- We can use *self-projective* internal potentials,

- Extended projections $\Pi_{prev(i)}^T$ can also be used, they project in all external ports' null space.

Such solutions are not yet implemented in our control; the internal dynamic is left open loop for the time being. However, we can tune the configuration through $B_a$.

### 4.3 PASSIVE VIRTUAL GUIDES

As stated before, we want to add the possibility to guide the movements of our virtual human (*guides* block of Figure 12). The most intuitive way to implement such guides is to project the error to be corrected by our operational space controller. We showed introducing the projection matrix, could break passivity. As in telerobotics [19], we used the virtual link concept, in order to realize passive projections. Following the mechanical analogy approach, passivity is ensured.

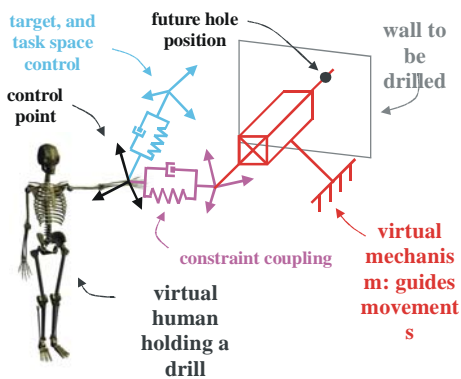

**Figure 14 Passively guided virtual human**

In Figure 14, the manikin holds a drill which axis is aligned with the future hole axis in the wall thanks to the simple virtual mechanism in red. Any other constraint could be expressed thanks to virtual mechanisms.

### 4.4 EXPERIMENTS AND RESULTS

The experiment consists in drilling a hole in a wall thanks to a drill, while lighting the future hole location thanks to a hand light. Both tools are guided, thanks to our virtual mechanisms framework. The drill can only move along a fixed axis with a fixed orientation. This means that the controller leaves only one degree of freedom to the operator. The direction of the spotlight is also driven automatically (leaving the three degrees of freedom of the light's position to the operator). Figure 15 depicts the ideal axis in green (a) and actual axes are in red (b).

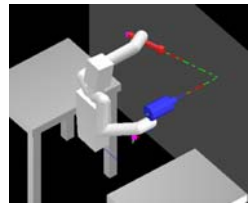 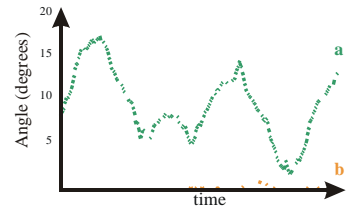

**Figure 15 Worker drilling a hole, guided by virtual mechanisms**   **Figure 16 Angle between ideal and actual axis of the drill, (a) without guide, and (b) with guide.**

In order to see the efficiency of our method, we drew the angle between the ideal axis, and the actual axis of the drill, as seen on Figure 16; in the case where the operator is completely free (green), and in case where the guide is on (orange). We also tested the collision engine. Figure 17 shows anti-collision in action.

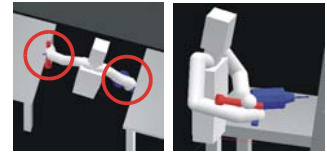 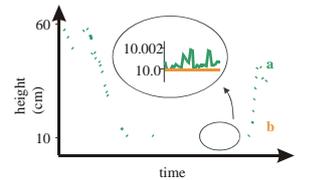

**Figure 17 Double, and self-collision**   **Figure 18 Hand (a), and obstacle's (b) height: no penetration.**

On the curve Figure 18, we can see the height of the table, which must not be penetrated (orange), and the height of the virtual human's hand (green), while reaching, and leaning on the table. We see that the hand never penetrates the table.

### 5. CONCLUSIONS

Two approaches have been presented to animate virtual mannequins. Indeed, the human tasks can be analysed with this both methods. The first one can be made when we have only the DMU but, nowadays, it not provides a global evaluation of the human fatigue. Some indices coming from ergonomic studies are already implemented in software but the results are partial. The other one can be made with human and the product. An ergonomist analyses the human motion to qualify its fatigue by using several methods. Such results will be made in future works and implemented in the context of aircraft industry.

## 6. ACKNOWLEDGMENTS